\documentclass[sigconf,nonacm]{acmart}
\setcopyright{none}
\usepackage{algorithm}
\usepackage{algpseudocode}
\usepackage{ragged2e}
\usepackage{balance}
\usepackage{titlesec}

\titlespacing\section{0pt}{12pt plus 2pt minus 2pt}{6pt plus 2pt minus 2pt}
\titlespacing\subsection{0pt}{10pt plus 2pt minus 2pt}{5pt plus 2pt minus 2pt}
\titlespacing\subsubsection{0pt}{8pt plus 1pt minus 1pt}{4pt plus 1pt minus 1pt}

\AtBeginDocument{%
  }
\begin{document}
\settopmatter{printacmref=false} 
\pagestyle{empty}

\title{Room Scene Discovery and Grouping in Unstructured Vacation Rental Image Collections}

\author{Vignesh Ram Nithin Kappagantula}
\email{vkappagantula@expediagroup.com}

\affiliation{%
  \institution{Expedia Group}
  \city{Seattle}
  \state{Washington}
  \country{USA}
}

\author{Shayan Hassantabar}
\email{shassantabar@expediagroup.com}

\affiliation{%
 \institution{Expedia Group}
 \city{Seattle}
 \state{Washington}
 \country{USA}}

\renewcommand{\shortauthors}{Kappagantula et al.}

\begin{abstract}
The rapid growth of vacation rental (VR) platforms has led to an increasing volume of property images, often uploaded without structured categorization. This lack of organization poses significant challenges for travelers attempting to understand the spatial layout of a property, particularly when multiple rooms of the same type are present. To address this issue, we introduce an effective approach for solving the room scene discovery and grouping problem, as well as identifying bed types within each bedroom group. This grouping is valuable for travelers to comprehend the spatial organization, layout, and the sleeping configuration of the property. We propose a computationally efficient machine learning pipeline characterized by low latency and the ability to perform effectively with sample-efficient learning, making it well-suited for real-time and data-scarce environments. The pipeline integrates a supervised room-type detection model, a supervised overlap detection model to identify the overlap similarity between two images, and a clustering algorithm to group the images of the same space together using the similarity scores. Additionally, the pipeline maps each bedroom group to the corresponding bed types specified in the property's metadata, based on the visual content present in the group's images using a Multi-modal Large Language Model (MLLM) \cite{inproceedings, ye2023mplug, liu2024improved} model. We evaluate the aforementioned models individually and also assess the pipeline in its entirety, observing strong performance that significantly outperforms established approaches such as contrastive learning and clustering with pretrained embeddings.
\end{abstract}
\renewcommand\footnotetextcopyrightpermission[1]{}

\maketitle

\section{Introduction}
With the growing popularity of vacation rental (VR) platforms, managing and organizing large collections of property images has become a significant challenge. Vacation rental listings often contain numerous images showcasing different rooms (e.g., bedroom, kitchen, living room, bathroom), but these images are typically uploaded without a structured categorization. For travelers, understanding the layout of a rental property is crucial for making informed booking decisions, as it helps them visualize the space and assess whether it meets their needs. However, without proper categorization, the navigation of unstructured images can be overwhelming, time-consuming, and, in some cases, confusing. Moreover, many properties contain multiple rooms of the same type (e.g., multiple bedrooms), making it even more difficult to differentiate spaces. Properly grouping images by specific room type and distinguishing between multiple rooms of the same type can improve the user experience and provide a clear spatial understanding of the property. In addition, identifying the type of bed associated with each bedroom in the property would give travelers a better understanding of the sleeping configurations that a property offers. Hence, providing a structured image catalog as well as the bed-type information can enhance the user experience, and assist them in the decision-making process. This can further drive business for the online VR platform.

Although critically important, there has been limited work to directly address this problem. A trivial approach to solve this problem is using feature representations extracted from the set of images, followed by unsupervised clustering to group them together based on similarity. 
Such representations from the images can be obtained using pre-trained visual encoders, such as \cite{9710580},  \cite{726791}, \cite{Dosovitskiy2020AnII}, \cite{tan2020efficientnetrethinkingmodelscaling}. Image encoders convert images into compact feature representations that capture essential visual information. Models like CNNs\cite{726791}  or vision transformers\cite{Dosovitskiy2020AnII} are commonly used for tasks such as classification, detection, or captioning. Although these representations are powerful in capturing the semantic information, applying clustering algorithms directly to these representations would not yield optimal room-scene groupings. The pre-trained representations primarily capture contextual similarities rather than fine-grained differences between rooms. As a result, images from different rooms with similar overall contexts may be clustered together. 
Contrastive learning \cite{pmlr-v119-chen20j}, \cite{NEURIPS2020_d89a66c7} using the supervision of room grouping data can learn close representations of images from the same rooms, allowing effective clustering. 
However, practically this process requires prohibitively large annotated datasets, which defeats the original purpose. In addition, contrastive learning-based methods often focus on broader categories (e.g., indoor vs. outdoor scenes) rather than the fine-grained specific room-level grouping needed for VR property images. Another challenge is that even with a given clustering, detecting the type of bed can still be challenging. Detecting the bed-type depends on a set of images, where each image may provide partial information about the bed type. As a result, the challenge still remains on how to efficiently use the structured grouping of bedroom images to detect 
the bed type. 

{\sloppy
To address the aforementioned challenges, we propose a novel framework comprising four sequential steps: i) room-type classification: classifying each image of the property to a specific room type, ii) pairwise image overlap detector: estimating the degree of similarity or overlap between a pair of images, iii) spectral clustering: a clustering-based algorithm that utilizes the overlap scores between all the combination of image pairs of a certain room type and groups the images of the same room space together, and iv) mapping clusters to textual entities: each bedroom cluster is mapped to the description of a specific bed type based on the images in the cluster.}
\interlinepenalty=0 \goodbreak 
We define room-type as either bedroom, living room, or bathroom images of the property. Since there are clear distinctions between the content in the images of each room-type, we first classify the images into the aforementioned categories. Hence, the first component of the pipeline is a trained classification model that identifies the room-type of an image. 
Doing so allows for the breakdown of the room grouping problem to grouping images of the same room type. This enables easier and more efficient processing later in the pipeline. To achieve this, we train a DINOv2 \cite{oquab2024dinov2learningrobustvisual} backbone classification model on the train domain image catalog to detect the room type. The trained model is highly capable of differentiating between the images that belong to these three room types.

Since two images from the same room might overlap with each other, we train a second model to identify the degree of overlap between pairs of images. 
To this end, a Siamese network model \cite{Koch2015SiameseNN} takes a pair of images as input and calculates the probability score representing the degree of overlap between the images. We estimate visual similarity across image pairs and use a similarity matrix to support clustering within room types. The challenge in training the Siamese network is the prohibitively costly process of collecting annotated image pairs.  To address this, we propose a training recipe that targets specific data augmentations to automate this process. Using self-supervised generated pairs significantly reduces the need for manually annotated pairs and improves the performance of the Siamese network model. 

The third component is a clustering module that groups the images of a room-type into specific rooms. We apply spectral clustering \cite{article} to the pre-computed pairwise overlap score matrix to determine the room groupings. 
We also make use of post-processing methods to reduce the noise and remove outliers in the final groupings. 

In the property on-boarding process, the owner of the property provides textual description of all the bed configurations in the property, for example, 1 King Bed, 2 Twin Beds. 
However, this information lacks the mapping to specific images of that property.
Hence, in the last step, we map the images of rooms containing beds to these textual descriptions.
To this end, we build a MLLM based on a pre-trained version of the Phi 3.5 \cite{abdin2024phi3technicalreporthighly} model.
We utilize Phi-3.5 for two main reasons. Firstly, the model is a small MLLM that is highly capable in image understanding. Secondly, it has the capability to take multiple images as input. 
We fine-tune this model on a dataset containing tuples of type (image cluster, text describing the bed configurations). 
The resulting model analyzes all images within a cluster to generate a textual description of the bed type. This process uses the available bed types from the property's metadata as a reference.

By systematically combining these components into a cohesive pipeline, our approach provides a robust framework for room scene discovery, grouping and bed-type identification, particularly for bedrooms. Our evaluations on a sample of 375 properties show an average normalized Adjusted Rand Index (ARI) score of 0.8065 and an average V-Measure score of 0.8284. We also use an optimization in the inference pipeline that leads to reduction in the inference time by $65.4\%$, resulting in much faster processing. The accuracy of the entire pipeline, which includes, grouping of images and mapping of bedroom image groups to its bed type, evaluated on a sample of 308 properties is 81.6\%. The accuracy improves by $36$\% when compared to the baseline solution of using feature vectors from pre-trained models to compute the overlap scores.

Evaluation on bed-type identification in bedrooms using exact string match accuracy shows an accuracy of $89\%$ on the validation dataset and $78\%$ on the test dataset. Considering the potential presence of noise in the predicted clusters, we achieve an accuracy of $78\%$ on the test dataset of $907$ image groups containing $3543$ images in total.

The major contributions of this work can be summarized as follows:
\begin{itemize}
    \item Propose a novel pipeline designed to efficiently address the problem of room scene discovery and grouping, as well as the identification of bed types in the bedrooms in a scalable manner.
    \item Introduce an innovative training strategy designed to train the pipeline with a minimal number of manually annotated examples, improving sample efficiency.
    \item Conduct extensive evaluations of each step of the proposed framework and showing the high performance of each component separately, and the overall end-to-end system. 
\end{itemize}

The remainder of this paper is structured as follows: Section \ref{sect:related_work} reviews related work on the components used in the pipeline. Section \ref{sect:methodology} details the methodology, explaining the different components in the proposed framework, and the dataset curation process for both pre-training and fine-tuning stages. 
Section \ref{sect:results} discusses the evaluation process and the metrics used to evaluate different components. 
Finally, Section \ref{sect:conclusion} concludes the paper.

\section{Related Work}
\label{sect:related_work}
In this Section,  we review some of the previous work in three
related areas: contrastive learning, Siamese networks and multi-modal LLMs. 

\subsection{Contrastive Learning}
Contrastive learning has emerged as a powerful approach in self-supervised representation learning, enabling models to learn from unlabelled data by maximizing agreement between similar pairs while minimizing it between dissimilar ones.
Notable frameworks such as SimCLR \cite{pmlr-v119-chen20j}, MoCo \cite{9157636}, and BYOL \cite{NEURIPS2020_f3ada80d} have demonstrated that contrastive objectives can produce rich, transferable image embeddings. These methods typically rely on instance discrimination, data augmentation, and large batch sizes to ensure diverse negative samples. However, they often lack explicit mechanisms to model pairwise relationships at a finer granularity, which can be crucial in clustering or grouping similar instances. Additionally, while contrastive learning excels at creating separable embeddings, it does not inherently produce interpretable or semantically grouped representations unless it is paired with downstream algorithms like clustering or supervised classifiers.

\subsection{Siamese Networks}
Siamese networks have emerged as a robust framework for learning embeddings by comparing input pairs through shared-parameter subnetworks, optimized with a distance-based loss. 
Initially proposed in \cite{bromley1993signature} for signature verification, these architectures explicitly optimize for proximity in embedding space, distinguishing them from global feature extractors and making them particularly effective for tasks requiring fine-grained similarity, such as face verification \cite{schroff2015facenet}, signature recognition, and few-shot classification \cite{Koch2015SiameseNN}. 

In a typical Siamese setup, two input instances are processed by identical subnetworks, and their outputs are compared using a distance function such as Euclidean distance or cosine similarity. 
Training uses loss functions such as N-pair loss \cite{sohn2016improved} or triplet loss \cite{schroff2015facenet}, which encourage embeddings of similar pairs to be close and dissimilar pairs to be far apart in the latent space. This enhances the performance on downstream applications such as clustering. Siamese networks are especially useful in few-shot learning, where only a few labeled examples are available. Instead of trying to classify images directly, these networks learn to measure how similar two images are. During inference, a new input is compared with the learned representations of a limited set of labeled examples. The network assigns a class label to the input based on its computed similarity to these reference examples. In our framework, we leverage Siamese networks as a classification models to detect the degree of overlap between a pair of images from the same room type. 

\subsection{Multi-Modal Large Language Models}
MLLMs have demonstrated the natural instruction-following and visual reasoning capabilities \cite{NEURIPS2023_6dcf277e, ye2023mplug, liu2024improved}. 
Phi-3.5-Vision \cite{abdin2024phi3technicalreporthighly} is a multimodal generative model with 4.2B parameters, capable of processing image or multi-image inputs alongside textual prompts and produce textual outputs. It comprises a CLIP ViT-L/14 image encoder \cite{pmlr-v119-chen20j} and a phi-3.5-mini transformer decoder. Visual tokens extracted from input images are interleaved with text tokens in an unordered manner. For multi-image scenarios, tokens from all images are simply concatenated.

The model undergoes pre-training on $0.5$ trillion tokens sourced from diverse datasets, including interleaved image-text documents \cite{li2024llavanextinterleavetacklingmultiimagevideo}, image-text pairs from FLD-5B \cite{inproceedings}, OCR-derived synthetic data, chart/table datasets, and text-only corpora. 
The training objective predicts the next text token, ignoring losses on image tokens. Post-training includes supervised fine-tuning (SFT) on 33B multimodal tokens covering tasks such as natural image reasoning, chart and diagram understanding. This is followed by Direct Preference Optimization (DPO) using both text-only and multi-modal preference data. 
These stages jointly enhance multi-modal reasoning while preserving the model’s core language generation capabilities. In our framework, we use a pretrained Phi-3.5 to build a model to generate the textual description of the bed configurations from the set of images from each bedroom.

\section{Methodology}
\label{sect:methodology}
In this section, we present the proposed room scene discovery and grouping methodology. First, we give a formal definition of the problem and provide an overview of the proposed methodology. Sections \ref{sect:room-type-classification} through \ref{sect:inference_time_optimization} discuss various parts of the proposed framework. 

\subsection{Framework Overview}
Our framing of the room scene discovery and grouping includes two main components, namely: (i) image grouping, where given all images of the property, the objective is to determine the type of the room depicted in each image and organize images of the same room spaces into separate groups, and (ii) bed configuration mapping: given the grouped images of each \emph{bedroom} to specify the bed-type in the room. 

\begin{figure*}[h!]
\includegraphics[height=10cm, width=\linewidth, keepaspectratio]{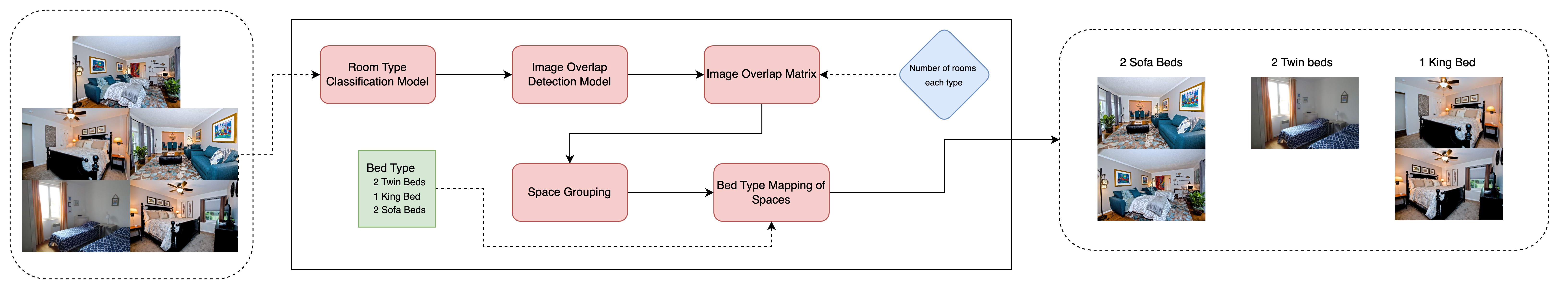} 
\caption{Overview of room scene discovery and grouping framework.}
\label{fig:Indoor room scene grouping overview}
\end{figure*}

Fig.~\ref{fig:Indoor room scene grouping overview} shows the overview of our proposed framework. First, to reduce the problem to grouping images of the same room type, we use a trained image classification model to identify the type of room shown in the image among the bedroom, living room, and bathroom. Section~\ref{sect:room-type-classification} explains this process. In the next step, an image overlap detection model predicts the degree of overlap between all possible pairs of images of the same room type. This process results in an overlap score matrix. The process of pairwise image overlap detection is explained in detail in Section~\ref{sect:Pairwise-Image-Overlap-Detection}. Using property metadata that specifies the number of room spaces of each type, a spectral clustering algorithm is applied to the image overlap matrix which groups the images according to the room spaces. The process of space grouping is discussed in detail in Section~\ref{sect:space-grouping}. The metadata provided by the property owners also provides a textual description of the various bed configurations offered in the property. Using a fine-tuned multi-modal LLM, we map each group of bedroom images to the corresponding bed configurations. Section~\ref{sect:space-mapping} further explains this process. 

\subsection{Room Type Classification}
\label{sect:room-type-classification}
Room type detection streamlines the process by grouping all images with similar characteristics into distinct room-type categories. This approach simplifies the algorithm by avoiding the increased time complexity that could result from processing all property images at once.

\begin{table*}
  \caption{Rules applied to the predictions of the multi-head classification model for identification of room type}
  \label{tab:rules_scenes}
  \begin{tabular}{cccccc}
    \toprule
    \textbf{Room}     & \textbf{Scenes} & \textbf{Concepts} & \textbf{Objects} & \textbf{Exclude Concepts} & \textbf{Exclude Objects}\\ \\
    \midrule
        Bathroom            & Bathroom               &             &    &  &          \\
        Living Room            & Guestroom, Property Interior, Undetermined             & Indoor         & Couch & Closeup & Bed           \\
        Bedroom            &    Guestroom, Property Interior, Undetermined            & Indoor            & Bed & Closeup &              \\
    \bottomrule
  \end{tabular}
\end{table*}

To identify the room types of all the property images, we train a multi-head classification model with pretrained DINOv2 encoder as the backbone network. This model includes three classification heads that enable the simultaneous prediction of different aspects (room scene, concept, and objects present) of the input image through independent output heads. A scene refers to the main setting shown in the image, such as a guestroom, bathroom, or pool. In contrast, concepts represent broader travel-related characteristics, such as whether the setting is indoors or outdoors. The rules in Table~\ref{tab:rules_scenes} are imposed on the outputs of the multi-head classification model to identify the room type of the image by leveraging object co-occurrence, the overall context represented by the image, and scene attributes. This model is trained especially on the travel domain data and taxonomy. It demonstrates high performance in detecting the tags related to the travel domain. 
\subsection{Pairwise Image Overlap Detection}
\label{sect:Pairwise-Image-Overlap-Detection}
In this section, we explain the process of curating the dataset  for pairwise image overlap detection for the sample-efficient training of the Siamese network model. 
\subsubsection{Siamese networks}
Since images of the same room space are captured at different angles, some images may overlap with each other. However, the two images of the same room might not always have an overlap if they are taken from different angles. In such cases, it is possible that a third image might have an overlap with both, enabling us to link the first two images together. Hence, processing and analyzing images in pairs enables us to detect these overlapping views and to understand the complete view of the room. 

To this end, we train a Siamese network model to determine whether a pair of images has an overlap in the view. This model takes two images as input and processes them individually to generate feature vectors for both images. We then combine these feature vectors using element-wise multiplication and pass the resulting combined feature vector through dense layers. 

The Siamese model predicts the overlap between all the pairs of images of a specific room type during inference, making low inference time important for scalability. Using larger backbone models negatively impacts scalability as the number of images of a specific room type increases. Therefore, we train the Siamese model with the EfficientnetV2-S \cite{tan2020efficientnetrethinkingmodelscaling} backbone as a binary classification model with a sigmoid focal loss\cite{8417976} objective to predict the presence of overlap in the view of any two images. The EfficientNetV2-S model has approximately 21.5 million parameters and a model size of around 84 MB. Its optimized architecture offers a favorable trade-off between accuracy and computational efficiency, enabling faster inference compared to larger models.

\subsubsection{Dataset generation}
\label{sect:dataset_generation}
\begin{figure}[h!]
\centering
\includegraphics[width=\linewidth, keepaspectratio]{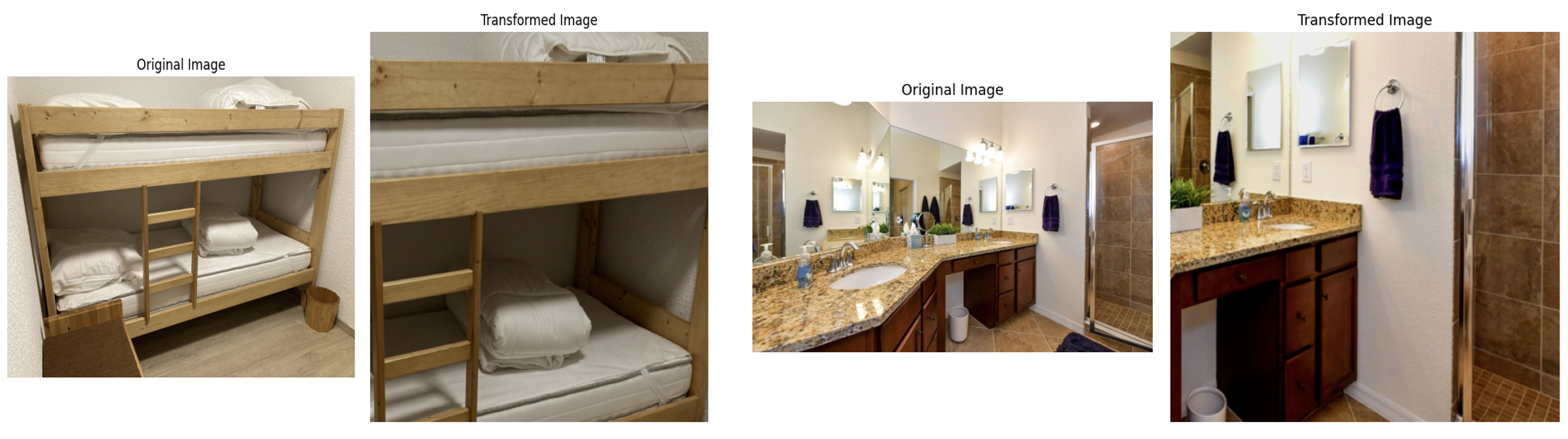}
\caption{Positive pairs obtained through data augmentation.}
\label{fig:self_supervised_pairs}
\end{figure}

\begin{figure}[h!]
\centering
\includegraphics[width=\linewidth, keepaspectratio]{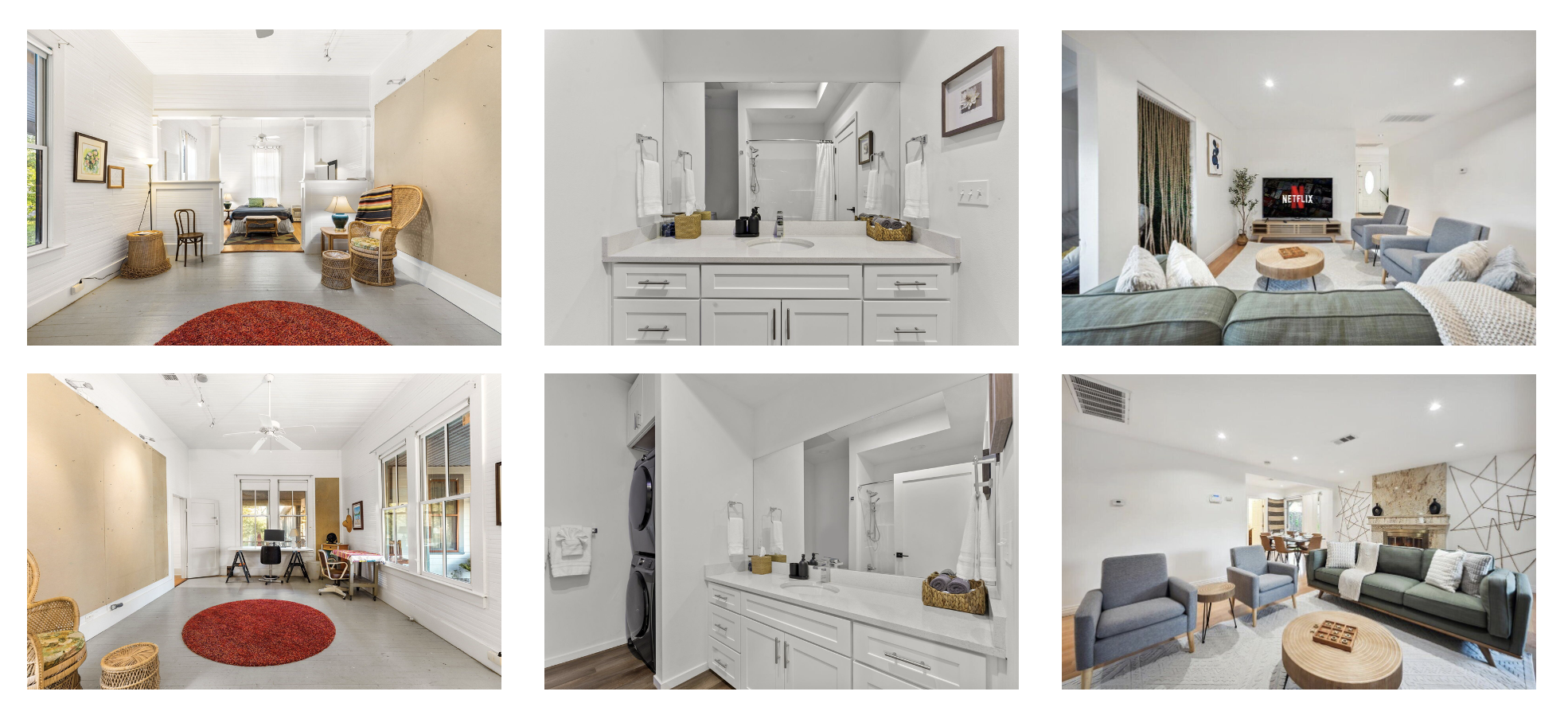}
\caption{Manually annotated challenging positive pairs. Every column of images is a pair.}
\label{fig:manually_annotated_pairs}
\end{figure}

Images from different room spaces within the same property can be challenging to distinguish due to their potentially strong visual similarities. This would require training the model with a well curated dataset of image pairs with overlap (or positive pairs) and image pairs with no overlap (or negative pairs). In addition, we have the option to create the positive pairs either manually or by self-supervision. To design a sample-efficient training recipe, we divide the dataset into two sources, (i) self-supervised pairs, and (ii) manually annotated pairs. 
\begin{itemize}
\item Self-supervised pairs - These image pairs result from applying data augmentation techniques to the original images in the dataset as shown in Fig.~\ref{fig:self_supervised_pairs}. 
It generates multiple views of the same image, enabling the model to learn useful representations without explicit labels. 
Hence, the original image will be any image from the property, the second image of the pair will result from applying data augmentation techniques to the first image. 
This simulates the process of manually annotating the image pairs. 

\item Manually annotated pairs : These image pairs enable the model to learn from challenging examples as shown in Fig.~\ref{fig:manually_annotated_pairs} that cannot be effectively simulated using traditional data augmentation techniques, particularly when attempting to replicate images resembling those in the original distribution. These are mostly image pairs with very little overlap and images taken from different angles to showcase the opposite sides of the room.
\end{itemize}
\vspace{-6pt}

\begin{figure}[h!]
\centering
\includegraphics[width=\linewidth, keepaspectratio]{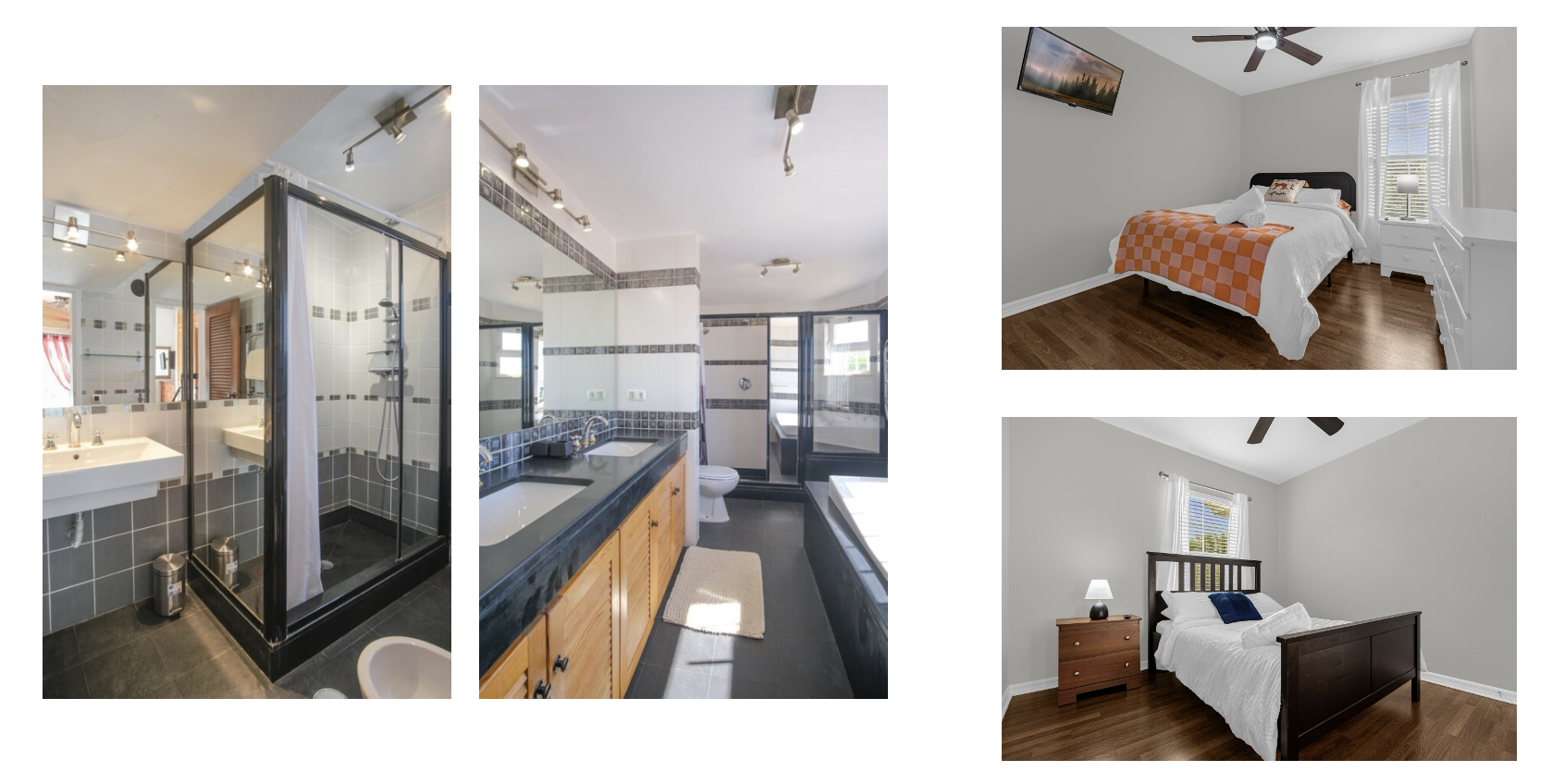}
\caption{Negative pairs.}
\label{fig:negative_pairs}
\end{figure}

Additionally, it is necessary to create negative image pairs dataset to help the model identify pairs without overlap even when the semantic context is quite similar. A random image from a property paired with another random image of the same room type but from a different room space within the property, as shown in Fig.~\ref{fig:negative_pairs} will form a negative pair. These negatives will help the model to learn the distinctions between images without overlap.

\subsubsection{Model training}
We train the Siamese network model using the image pairs from the dataset generated in Section \ref{sect:dataset_generation}. 
To effectively train the model with limited number of manually annotated dataset, the training is done in two steps as shown in Alg.~\ref{alg:siamese_pretraining_finetuning}

\begin{itemize}
    \item Pre-training : Data augmentation and random selection of two images from different rooms of the same property easily produce positive and negative pairs respectively. We generate approximately 100,000 image pairs for both positive and negative pairs, creating a dataset to pre-train the Siamese network. An equal number of positive and negative pairs were used to ensure balanced learning, preventing class imbalance bias and enabling the model to learn both classes effectively. This pretraining will help the model learn general features of patterns that can boost the performance of the task, especially when the labeled data is limited. Pretraining with these pairs enables the model to learn to identify image pairs with slight camera angle changes, which data augmentation simulates. It also helps the model recognize image pairs with no overlap, as it trains on numerous such negative examples.
    \item Finetuning :  The finetuning of the Siamese network model enables to identify the positive pairs or the images with overlap from the same room. The model especially learns to identify the positive image pairs which the data augmentation technique failed to simulate on the original image as shown in Fig.\ref{fig:manually_annotated_pairs}. We annotated approximately 3500 image pairs manually. Since there are limited number of manually annotated positive pairs when compared to the self-supervised pairs, we finetune the Siamese network model with this limited data. 
\end{itemize}
 With the training strategy split into two stages, the model trains to identify the image pairs with no overlap and maximum overlap during the pre-training stage. And it trains to identify the image pairs that have overlap but are very difficult to simulate by data augmentation during the fine-tuning stage. In addition, the fine-tuning dataset contains the same number of self-supervised positive pairs and negative pairs equal to the number of manually annotated pairs. This is done to prevent the model from catastrophic forgetting from the pre-training stage. 

\begin{algorithm}[t]
\caption{Siamese Network Pretraining and Finetuning}
\label{alg:siamese_pretraining_finetuning}
\begin{algorithmic}[1]
\Require 
    $\mathcal{D}_{\text{pos-self}}$: self-supervised positive pairs, \\
    $\mathcal{D}_{\text{pos-manual}}$: manually annotated positive pairs, \\
    $\mathcal{D}_{\text{neg}}$: negative pairs
\Ensure Fine-tuned Siamese model $M$
\State \textbf{Construct Datasets:}
\State $\mathcal{D}_{\text{pre}} \gets \{ \mathcal{D}_{\text{pos-self}}, \mathcal{D}_{\text{neg}} \}$
\State $\mathcal{D}_{\text{fine}} \gets \{ \mathcal{D}_{\text{pos-manual}}, \mathcal{D}_{\text{pos-self}}, \mathcal{D}_{\text{neg}} \}$
\State \textbf{Set Hyperparameters:}
\State $\eta_{\text{pre}} = 10 \times \eta_{\text{fine}}$
\State Number of frozen layers: $L_{\text{pre}}$ (pretraining), $L_{\text{fine}}$ (finetuning), where $L_{\text{fine}} > L_{\text{pre}}$
\State \textbf{Pretraining Phase:}
\State Initialize model $M$ with parameters $\Theta$
\State Freeze first $L_{\text{pre}}$ layers of $M$
\ForAll{minibatch $(\text{img}_1, \text{img}_2, y) \in \mathcal{D}_{\text{pre}}$}
    \State Compute loss $\mathcal{L}_{\text{pre}}$
    \State Update $\Theta$ using optimizer with learning rate $\eta_{\text{pre}}$
\EndFor
\State \textbf{Finetuning Phase:}
\State Load pretrained parameters into $M$
\State Freeze first $L_{\text{fine}}$ layers of $M$
\ForAll{minibatch $(\text{img}_1, \text{img}_2, y) \in \mathcal{D}_{\text{fine}}$}
    \State Compute loss $\mathcal{L}_{\text{fine}}$
    \State Update $\Theta$ using optimizer with learning rate $\eta_{\text{fine}}$
\EndFor
\end{algorithmic}
\end{algorithm}

\subsection{Space Grouping}
\label{sect:space-grouping}
This section explains the methodology for using the trained Siamese network model to generate overlap similarity matrix for each room type and apply spectral clustering algorithm on top of the matrix to obtain image groupings of the room spaces.

\subsubsection{Room overlap score matrix}
\begin{figure}[h!]
\centering
\includegraphics[width=\linewidth, keepaspectratio]{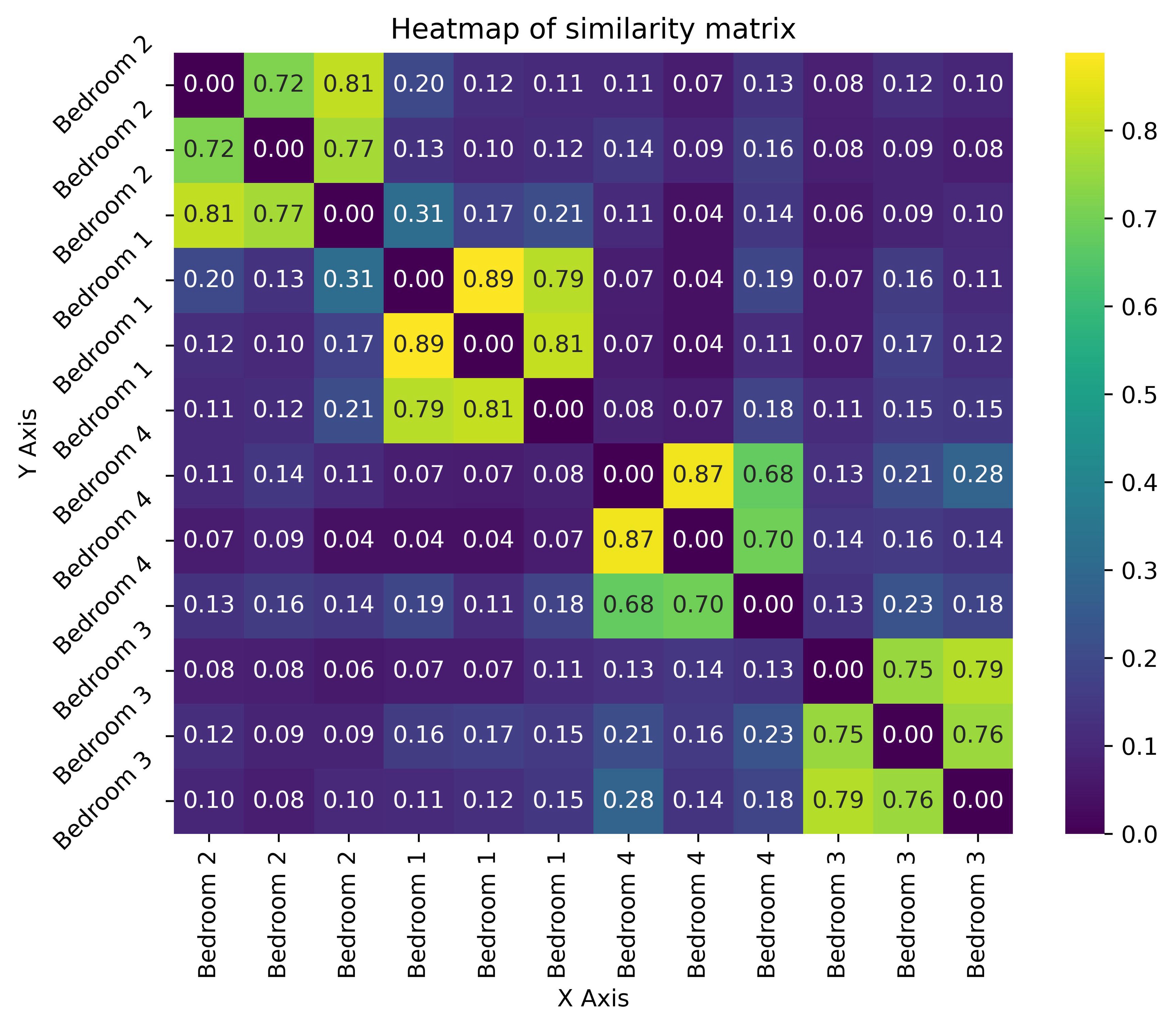}
\caption{Overlap score matrix of bedroom images in a property. In this example, the property has 4 bedrooms and a total of 12 bedroom images forming a 12x12 matrix.}
\label{fig:overlap_score_matrix}
\end{figure}

To group images of a room type into their respective room spaces, we predict the overlap score between all possible image pairs using the Siamese network model from Section \ref{sect:Pairwise-Image-Overlap-Detection}.
This results in an overlap score matrix as shown in Fig. \ref{fig:overlap_score_matrix}. Similarly, we obtain the overlap score matrix for all the room types which the DINOv2 multi-headed scene classification model identifies.

\subsubsection{Spectral clustering}
The spectral clustering algorithm uses the pairwise overlap score matrix of each room type to group images into distinct room spaces. 
By leveraging eigenvectors of the similarity matrix, spectral clustering captures complex, nonlinear relationships and efficiently clusters the data using k-means in a transformed space identifying the boundaries between different rooms of the same type. 

Clustering is an unsupervised learning technique used to group similar data points without prior knowledge of the labels. However, in the process of grouping images into their respective spaces, the images might not be clearly separable, which can lead to ambiguous cluster boundaries. This lack of distinct separation often results in outlier images that don’t fit well into any cluster and being incorrectly grouped. These outliers can reduce the overall accuracy and interpretability, impacting the quality and usefulness of the room grouping results. To reduce noise or remove outliers in the image clusters, the mean overlap score of each image with other images in the cluster is calculated, and images with a mean similarity score below a percentage of the maximum mean similarity are removed. This process decreases the number of images in the cluster, minimizing noise and improving the cluster's precision.

\subsection{Space Mapping to Textual Descriptions}
\label{sect:space-mapping}
This section describes the methodology for identifying the type of bed associated with the cluster of images of the room space as shown in Alg. \ref{alg:space_mapping}.

\subsubsection{Training dataset}
Properties listed on the vacation rental platforms feature various bed types in each bedroom. To provide travelers with a clear property layout, identifying the bed types in different bedrooms offers valuable insight into the sleeping configuration. Since we group bedroom images into clusters, we map each cluster to the corresponding bed type present in those images. The sourcing of the dataset for training a model to identify the bed type is done from select properties where metadata is available for clustered bedroom spaces along with their bed types. We curated a total of 16937 bedroom groups, containing about 65936 images and corresponding bed type labels, as the training dataset. Similarly, we also curated a validation dataset, comprising 907 bedroom groups with 3543 images. Both the training and validation datasets consist of image groups as input, accompanied by a prompt describing the task and a corresponding bed type label, such as "1 King Bed" or "1 Queen Bed". The prompt trains the model to generate the type of bed based on the list of bed types available in the property's metadata.

\subsubsection{Space mapping}
Since multiple images of a room space are grouped into a cluster, we need to analyze all the images simultaneously to determine the type of bed in the bedroom. To this end, we fine-tune the Phi-3.5 MLLM as it has the capability to take multiple images as input. We use the LoRA \cite{hu2021loralowrankadaptationlarge} technique to fine-tune the model, allowing it to efficiently adapt to the task of identifying the type of bed while maintaining computational efficiency. To enhance the model’s ability to accurately identify bed types from room-space images, we apply LoRA specifically to key linear layers responsible for both language and vision processing. By fine-tuning these specific layers, the model learns to extract meaningful visual features related to the identification of bed types, such as bed size, shape, and arrangement within the room. This enables the model to accurately determine the bed type from a given set of options in property metadata, ensuring precise mapping of clustered bedroom images.

During the inference process, the model maps the image clusters of bedrooms to the predefined bed types listed in the property's metadata as shown in Alg. \ref{alg:space_mapping}. To ensure a one-to-one mapping and avoid multiple clusters being assigned to the same bed type, the image clusters are processed sequentially. The model iteratively matches image clusters to bed types defined in the metadata, subject to coverage constraints.

\begin{algorithm}[t]
\caption{Space Mapping Algorithm for Inference}
\label{alg:space_mapping}
\begin{algorithmic}[1]
\Require
    Set of image groupings $\{G_1, G_2, \dots, G_n\}$ for $n$ bedrooms, Bed types list $B$ from property metadata
\Ensure
    Mapped bed types $\{B_1, B_2, \dots, B_n\}$ corresponding to each $G_i$

\State Construct frequency dictionary $D$ from $B$. $D[B_k]$ stores the count of bed type $B_k$.

\For{$i = 1$ to $n$}
    \State Let $G_i = \{ \text{img}_1, \text{img}_2, \dots \}$ be the $i^{th}$ group
    \State Extract candidate bed types $O \gets \text{keys}(D)$
    \State Construct prompt including $G_i$ and options $O$
    \State Query model with the prompt to predict bed type $B_k \in O$
    \State Assign group $G_i \gets B_k$ bedtype
    \State Update frequency dictionary: $D[B_k] \gets D[B_k] - 1$
    \If{$D[B_k] = 0$}
        \State Remove $B_k$ from $D$
    \EndIf
\EndFor
\end{algorithmic}
\end{algorithm}

\begin{table*}
  \caption{Room scene grouping metrics}
  \label{tab:property_clustering}
  \centering
  \begin{tabular}{cccccc}
    \toprule
    \multicolumn{2}{c}{\textbf{}} & \multicolumn{2}{c}{\textbf{Before noise removal}} & \multicolumn{2}{c}{\textbf{After noise removal}} \\
    \cmidrule(lr){1-2} \cmidrule(lr){3-4} \cmidrule(lr){5-6}
    \textbf{Number of bedrooms} & \textbf{Number of properties} & \textbf{ARI normalized} & \textbf{V-measure} & \textbf{ARI Normalized} & \textbf{V-measure} \\
    \midrule
    2 & 99 & 0.8727  &  0.8311   & 0.8762   & 0.855   \\
    3 & 101 & 0.8682  & 0.8752  & 0.8718   & 0.886   \\
    4 & 94 & 0.8502  & 0.889   & 0.8568   & 0.900   \\
    more than 4 & 81 & 0.8221  & 0.885   & 0.8305   & 0.897   \\
    \bottomrule
  \end{tabular}
\end{table*}

\subsection{Inference Time Optimizations}
\label{sect:inference_time_optimization}
This section details the optimization techniques that reduce inference time and enhance the scalability of the overall pipeline. We applied two optimizations to the inference pipeline.
\begin{itemize}
\item 
Precomputing the multi-headed classification model's inference results for all hotels and vacation rentals listed in the vacation rental catalog, allowing for efficient and faster retrieval during real-time processing. We leverage these precomputed outputs on the fly to apply additional rules mentioned in Table.~\ref{tab:rules_scenes} to the predictions of every image, to identify the specific room type of the image. This approach minimizes overall inference time, improves pipeline scalability.
\item 
During inference, to group the images of specific room type into their respective room spaces, we compute the image overlap score between all the images pairs of that room using the trained Siamese network model. If there are n images of a specific room type, the total number of possible image pairs is $\binom{n}{2}$.Since each pair requires two forward propagations of the Siamese network model to compute the score, the total number of forward propagations needed would be $2*\binom{n}{2}$ which is a significantly large number. To minimize the excessive forward propagation required to compute the image overlap score matrix for a room, we divide the Siamese Network into a feature encoder and a classification head. By passing each image through the feature encoder once, we obtain its feature vector, eliminating redundant computations. The classification head determines the score between any pair of images by processing their corresponding precomputed feature vectors, which significantly reduces computational overhead.
\end{itemize}
By applying these two optimizations to the inference pipeline, we achieve improved efficiency, reduced latency, and enhanced overall performance. The first optimization streamlines computational overhead, while the second further refines resource utilization, leading to a more optimized and scalable solution. These enhancements improve run-time performance and indicate readiness for scalable testing in applied environments.

\section{Results}
\label{sect:results}
In this section, we present the results of the models trained and used in this pipeline. Our pipeline consists of a room-type classification model to classify the images of the properties to a specific room, an overlap detection model which takes two images at a time to provide the overlap score between the two images, a clustering algorithm to group the images of the same room type using the overlap score matrix, and a vision language model to identify the bed type present in the group of images obtained from the clustering algorithm. We also present the quantitative results of the inference-time optimizations explained in Section \ref{sect:inference_time_optimization}.   

\subsection{Ablation Study}
\subsubsection{Scene classification}
We evaluated two models, a multi-head classification model with DINOv2 encoder trained on data specific to travel domain, and BLIP-2, a vision-language model on a test dataset of 25 properties which totally consists of 832 images. We observe that the trained classification model has better precision and F1-score for all categories than the BLIP-2 model, as shown in Table \ref{tab:dinov2_room_type_results} and \ref{tab:blip2_room_type_results}. Furthermore, for most classes, the BLIP-2 model achieves marginally higher recall compared to the DINOv2 model.

\subsubsection{Overlap detection - Pretraining and Finetuning results}
We evaluate the performance of the overlap detection model using a manually annotated dataset of 1898 image pairs with 1000 positive pairs and 898 negative pairs as shown in Table~\ref{tab:Siamese_overlap_detection}.
Our approach leverages the self-supervised positive pairs and negative pairs to pre-train the Siamese network model. We used considerably fewer manually annotated pairs, self-supervised pairs, and negative pairs to fine-tune the model. This improved the recall and F1-score by 44.6\% and 18.5\% respectively when compared to the pre-trained model.

The fine-tuning of the Siamese network model shows massive improvement in recall demonstrates that the model is better aligned with understanding the manually annotated pairs which include complex angles from the original property images not captured by the self-supervised dataset obtained via data augmentation. Fine-tuning also leads to a marginal drop in precision by $6$\% when compared to the precision of the pre-trained model due to more false positives. This may limit the effectiveness of the model's hard predictions. However, the soft probability scores still encode valuable information about the overlap similarities. This approach effectively utilizes the underlying latent structure to cluster using the overlap similarity scores, revealing insights that are not often apparent when utilizing embeddings obtained directly from the images. 
\begin{table}
  \caption{Metrics for DINOv2 room type classification}
  \label{tab:dinov2_room_type_results}
  \begin{tabular}{cccc}
    \toprule
    \textbf{Class} & \textbf{Precision} & \textbf{Recall} & \textbf{F1-Score} \\
    \midrule
    Bedroom & 0.95 & 0.98 & 0.97 \\
    Bathroom & 1 & 0.95 & 0.98 \\
    Living Room & 0.88 & 0.84 & 0.86 \\
    Other & 0.96 & 0.99 & 0.96 \\ 
    \bottomrule
  \end{tabular}
\end{table}

\begin{table}
  \caption{Metrics for Blip2 VLM room type classification}
  \label{tab:blip2_room_type_results}
  \begin{tabular}{cccc}
    \toprule
    \textbf{Class} & \textbf{Precision} & \textbf{Recall} & \textbf{F1-Score} \\
    \midrule
    Bedroom & 0.898 & 0.994 & 0.945 \\
    Bathroom & 0.902 & 0.965 & 0.93 \\
    Living Room & 0.818 & 0.882 & 0.85 \\
    Other & 0.978 & 0.912 & 0.942 \\ 
    \bottomrule
  \end{tabular}
\end{table}

\subsubsection{Clustering}
Table~\ref{tab:property_clustering} presents the performance metrics of the clustering between different groups of properties based on the number of bedrooms, evaluated before and after the noise removal. The removal of noise consistently improves clustering quality in all properties. For properties with two bedrooms, we observe an increase in normalized ARI from $0.8727$ to $0.8762$ after removing the noise and similar gains in V-measure from $0.8311$ to $0.855$. A comparable trend is seen in properties with more than 2 bedrooms, where post-cleaning metrics demonstrate better alignment with ground truth labels, suggesting that noise removal enhances cluster separability. The average normalized ARI increased from $0.8014$ to $0.8065$ and average V-measure increased from $0.8150$ to $0.8284$ after removing the noisy outliers. These results highlight that noise in the data significantly impacts clustering performance, and such post-processing steps helps in improving the quality of unsupervised learning outcomes. 

The group with more than four bedrooms has relatively lower ARI scores due to the complexity of the images and the increase in the number of images for clustering. Since V-measure is the harmonic mean of homogeneity and completeness, increasing the number of bedrooms in properties tends to improve homogeneity, as more clusters can better isolate distinct classes. However, completeness may not improve as significantly, since it depends on how effectively the pipeline groups images of the same bedroom class. As a result, the V-measure may initially rise but eventually reach a plateau as observed in Table \ref{tab:property_clustering}. 

\begin{table}
  \caption{Metrics for the finetuning of the Siamese overlap detection model when compared to the pretraining}
  \label{tab:Siamese_overlap_detection}
  \begin{tabular}{cccc}
    \toprule
    \textbf{Training stage} & \textbf{Precision} & \textbf{Recall} & \textbf{F1-Score} \\
    \midrule
    Finetuning & $0.94\times$ & $1.446\times$ & $1.185\times$ \\
    \bottomrule
  \end{tabular}
\end{table}

\subsubsection{Bed type space mapping}
We conducted inference experiments using the fine-tuned Phi-3.5 model to predict the bed type of images of a room. The initial experiment evaluates the model as a standalone component to assess its performance independently of the full pipeline. The evaluation uses $907$ groups containing a total of $3543$ images and achieves an average accuracy of $89\%$ in predicting the correct type of bed. This standalone evaluation presents the model with accurate image groups containing no incorrect images. However, during the actual inference of the pipeline, the image groups may occasionally include incorrect images due to error occurring in the earlier stages of the pipeline, such as room type classification, overlap similarity detection therefore achieving an accuracy of $78\%$ on the same set of $907$ image groups. 
 
\subsection{End-To-End Performance Evaluation}
To evaluate the performance of the entire pipeline, we selected a sample of 308 properties, processed them through the entire pipeline and the results were manually evaluated. The accuracy of the grouping of images according to the room scene type and the mapping of the image groups to the accurate bed type is 81.6\% and is improved by $36\%$ when compared to the baseline approach of clustering images based on features extracted by an image encoder. This points toward the superior performance of our proposed framework. 

\subsection{Inference Latency Optimizations}
The optimization of splitting the pairwise image overlap detector into a feature encoder and a classification head reduces the redundant computation of the feature vectors of the images for all the combination of image pairs. We cache the feature vectors of all images for repeated use. This optimization reduces the average inference time of the overlap detection model on all pairs of images by $65.4$\% when generating predictions for $200$ properties through the pipeline.

\section{Conclusion}
\label{sect:conclusion}
In this study, we developed a novel pipeline to automatically group indoor images of a vacation rental properties based on the specific room scenes they depict and identify the type of the bed present in the bedroom groups. This is very essential in understanding the layout of the property and identifying the bed type would provide a better understanding of the sleeping configuration of the vacation rental property. We proposed the integration of the soft labels produced by the Siamese overlap detection model with the spectral clustering algorithm to group the images, and we finetuned the Phi-3.5 MLLM to find the bed type in each bedroom group. The accuracy of the entire pipeline is 81.6\% and has improved by 36\% compared to the baseline approach of clustering images based on its features extracted by a pre-trained image encoder.

\newpage

\bibliographystyle{ACM-Reference-Format}
\bibliography{references}
\balance
\end{document}